\documentclass[runningheads]{llncs}
\usepackage{amsmath,amssymb,amsfonts,adjustbox}
\usepackage[T1]{fontenc}
\usepackage{graphicx}
\usepackage[absolute,overlay]{textpos}

\begin{document}

\begin{textblock*}{\textwidth}(0cm,0cm)
\noindent Please cite as follows:\\S. Filippou, A. Tsiartas, P. Hadjineophytou, S. Christofides, K. Malialis, C. G. Panayiotou. Improving Customer Experience in Call Centers with Intelligent Customer-Agent Pairing. In International Conference on Artificial Intelligence Applications and Innovations (AIAI), 2023.
\end{textblock*}

\title{Improving Customer Experience in Call Centers with Intelligent Customer-Agent Pairing}

\author{Stylianos Filippou\inst{1}\orcidID{0000-0001-8123-8309} \and
Andreas Tsiartas\inst{3} \and\\
Petros Hadjineophytou\inst{3} \and
Spyros Christofides\inst{3} \and\\
Kleanthis Malialis\inst{1}\orcidID{0000-0003-3432-7434} \and\\
Christos G. Panayiotou\inst{1,2}\orcidID{0000-0002-6476-9025}}

\authorrunning{S. Filippou, et al.}

\institute{
\textit{KIOS Research and Innovation Center of Excellence, University of Cyprus}\\\and
\textit{Department of Electrical and Computer Engineering, University of Cyprus}\\\and
\textit{Cyprus Telecommunications Authority (CYTA)}\\
\email{
\{filippou.stylianos, malialis.kleanthis, christop\}@ucy.ac.cy
}\\
\email{\{andreas.tsiartas, petros.hadjineophytou, spyros.christofides\}@cyta.com.cy}
}

\maketitle             

\begin{abstract}
Customer experience plays a critical role for a profitable organisation or company. A satisfied customer for a company corresponds to higher rates of customer retention, and better representation in the market. One way to improve customer experience is to optimize the functionality of its call center. In this work, we have collaborated with the largest provider of telecommunications and Internet access in the country, and we formulate the customer-agent pairing problem as a machine learning problem. The proposed learning-based method causes a significant improvement in performance of about $215\%$ compared to a rule-based method.

\end{abstract}

\section{Introduction}

Organisations or companies set high standards for providing excellent products and services to expand in the market, to retain current customers, and attract new ones. Customer experience  significantly affects the  loyalty  and  satisfaction  of  the  customer in  relation  to a company’s  products  and  services \cite{Konstantinos2009Data}. It is a top priority for any company or organisation, and it constitutes a vital component of its commercial and marketing strategy.

Nowadays, there exist multiple channels through which a customer can contact an organisation; one of the most widely used is the call center. A call center is a department within an organisation, that handles a large amount of incoming calls related to their products and services. The organisation through the call center collects and stores a variety of historical information through the different media of interaction with the customer. The main aim of the call center is to assist customers and answer any enquiry, therefore, a good functioning call center can drastically improve customer experience. By keeping customers satisfied, an organization can achieve their objectives, which among others, it includes customer retention through customer satisfaction \cite{Konstantinos2009Data}. Moreover, by successfully assisting customers, an overall positive experience can attract new customers.

An important way of improving customer experience through the call center is by minimizing the waiting time of a customer until an agent (i.e., call operator) assists them. Minimization of the call duration can be achieved by avoiding the traditional interaction via keypad, which redirects the customer after a series of keypad selections to the relevant agent who can assist with their enquiry. Minimizing each call duration offers significant benefits which are: (i) a customer is assisted faster; (ii) a company can assist more customers in the same amount of time; and (iii) a company saves valuable resources, e.g., by re-assigning agents to important problems.

The contributions of this work are as follows. We have collaborated with the Customer Support team of the largest provider of telecommunications and Internet access in the country, and have formulated the customer-agent pairing problem in their call center as a machine learning problem. We have conducted a rich experimental study using realistic data provided by the organisation, and examining various learning models. The proposed learning-based method is statistically 2.15 times better than a rule-based method.

The rest of the paper is organised as follows. Section~\ref{sec:related} reviews the related work. The problem formulation and proposed method are discussed in Section~\ref{sec:formulation}. Section~\ref{sec:setup} presents the datasets, classification models, and evaluation metrics used in this study. A description of the results and the comparative study are presented in Section~\ref{sec:results}. We conclude in Section~\ref{sec:conclusions}.

\section{Related Work}\label{sec:related}
Customer Relationship Management (CRM) is the strategy for building, managing and strengthening loyal and long-lasting customer relationships. According to \cite{Konstantinos2009Data} there are two main objectives of CRM. First, customer retention through customer satisfaction, and second customer development through customer insight. To achieve both objectives, organizations should focus on customer needs, behavior and preferences. Machine learning algorithms have played a major role towards achieving these objectives; many studies use such methods in CRM-related tasks.

\textbf{Customer identification and segmentation} is an important task for any organization because it can identify customer requirements and divide customers into groups using demographic data, such as age, location, gender, occupation etc. In \cite{wassouf2020Predictive} the authors compared various machine learning algorithms in order to group customers using these features: total call duration, frequency of using a service, and money spent during a certain period. In \cite{Singh2020Machine} they grouped customers into a number of classes using Naive Bayes, Decision Tree and MLP, while in \cite{kamthania2018Market} they used dimensionality reduction (PCA) and clustering (k-mode) to group customers.

\textbf{Customer attraction} is the task of attracting customers to an organization's products and services. In \cite{Martnez2020AML} the authors have applied machine learning to predict purchase of services, while in \cite{Singh2018Customer} the authors have used machine learning to identify prospect customers.

\textbf{Customer retention} refers to strategies that are, typically, targeted on customers that are most likely to abandon a service. According to \cite{Singh2021inclusive}, retention strategies should be applied by all organizations or companies as they are considered to be cost-effective (e.g., compared to attracting new customers). In \cite{Sahar2018Machine} the authors have applied various machine learning algorithms on a telecommunication dataset, while authors in \cite{Emtiyaz2012Customers} used a semi-supervised learning to retain their valuable customers.

\textbf{Customer churning} is the task of identifying the cost of losing customers. In \cite{Ahmad2019Customer}, the authors have applied machine learning for churn prediction in the area of telecommunications. Also in the same area, the authors in \cite{Mishra2017Novel} have proposed deep fully-connected and convolutional neural networks for churn prediction in the area of telecommunications.

\textbf{Customer lifetime value (CLV)} refers to the task of identifying the approaches that can create value to organizations, optimize their resources, and maximize their profits. In \cite{Chen2018Estimating}, the authors used tree-based learning algorithms to define CLV in airline travelers, and classified them as high, medium and low value travelers. In \cite{Salehinejad2016Customer} using features such as client loyalty number, recency, frequency and monetary, the authors proposed a customer shopping behavior model using recurrent neural networks.

\textbf{Customer-agent pairing} focuses on the successful communication between agents (i.e., call operators) and customers, thus maximizing customer satisfaction. To enhance the customer's call experience, organizations utilize historical and demographic data to improve the service to the customer by minimising the call duration, from the point a customer contacts the call center to the point the user enquiry is satisfied. To our knowledge, not many studies explored this problem. In \cite{Mehrbod2018Caller} the authors used biographic information and historical data to find the best pairing of callers and agents for the call center of an insurance company.
This work is closer in spirit to ours, however, the focus of our work is in the area of telecommunications. Lastly, in this study \cite{Golub2020Optimization}, the features that were considered are words, from speech to text conversion system available at the organisation. This method relies on an additional step where the customer is first required to describe the reason for calling, as opposed to interacting with their keypad which is what our proposed method is intended for.

\section{Intelligent Customer-Agent Pairing}\label{sec:formulation}

\subsection{Problem Formulation}
We consider a centralised call center that aims to provide services to existing or potential customers. The call center consists of many components, from which we will focus on the Interactive Voice Response (IVR) and the available assistance departments. The IVR component is a technology that allows a computer to interact with humans through input via a keypad. It provides an automated first contact with a customer before forwarding the call to the appropriate department. The IVR provides the user with a predefined number of options. Each initial category consists of a subgroup of options for the precise request identification, which will help the IVR to assign the caller to the queue of the appropriate department. Each department consists of a specialized group of personnel (the ``agents'' or ``operators'') that are qualified to help the customers with specific enquires. The overview of the current flow in IVR can be seen in Fig~\ref{fig:overviewSolution}. We describe Stages 1-2 here, while the prediction stage is described in the next section.

\textbf{Stage 1.} The customers voice call will be forwarded to the IVR system. At stage 1 the user will receive a welcome audio message, and be presented with the available $J$ options. The user will use the keypad to make their desired choice that fits most their enquiry. Depending on the selection, this process can be repeated. It is assumed that the average duration time of this stage is $t_{stage^1}$.

\textbf{Stage 2.} The user will be assigned to the corresponding $N$ queues, one for each department. Finally, the user is able to communicate with the appropriate personnel. The average duration for this stage is assumed to be $t_{stage^2}$.

The average time duration of any call $c$ at the call center is:
\begin{equation}\label{eq:userTime}
   T^{c} = t_{stage^1} + t_{stage^2}
\end{equation}

Let the total number of calls be $C$, the total average duration of all calls is:
\begin{equation}\label{eq:totalTime}
   T^{total} = \sum_{c=1}^C T^c
\end{equation}

\subsection{ML-based Pairing}
\begin{figure*}[t!]
	\centering
	\includegraphics[scale=0.25]{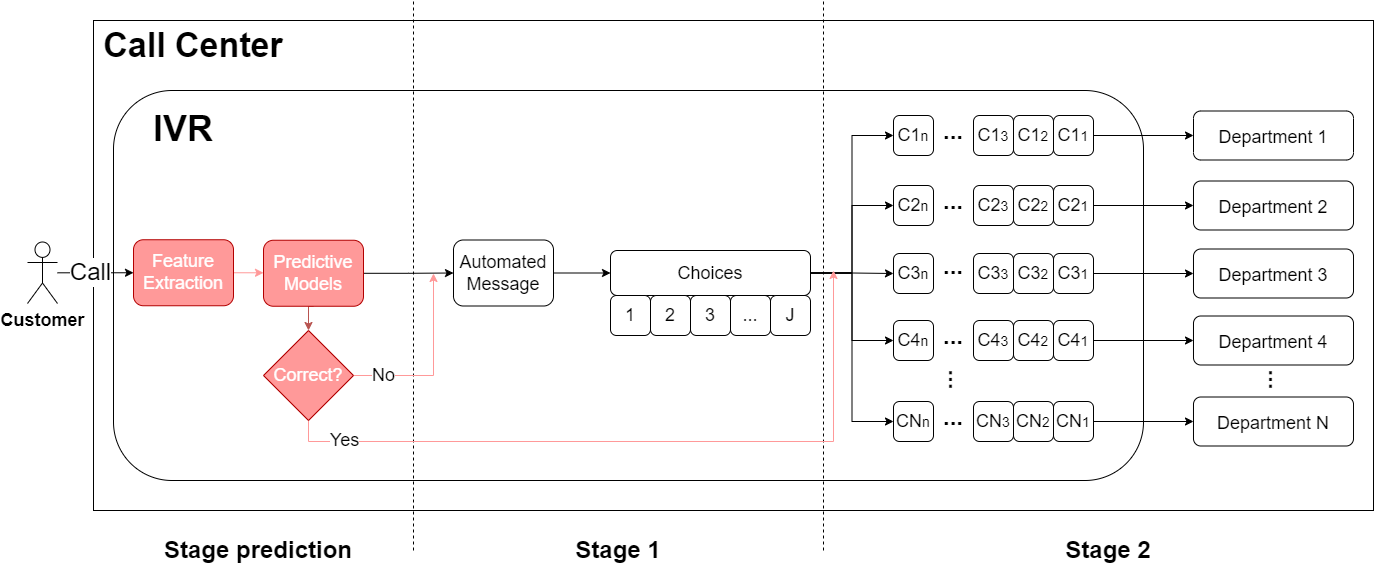}
	\caption{The proposed solution overview of the Customer's call procedure}
	\label{fig:overviewSolution}
\end{figure*}

Intelligent pairing can be achieved by classifying the call of each user before stage 1, and assigning the user to the correct queue department at stage 2. This is shown in red in Fig~\ref{fig:overviewSolution}. By achieving that, the user can skip the time consuming stage 1, and given a quick verification of the user they could directly forward the call to a department's queue, resulting in a faster service and an enhanced customer experience.

Let the identifier (e.g., telephone) of user $u$ that contacts the call center be $id(u) \in \mathbb{Z}^+$. A feature generating process $g: \mathbf{Z}^+ \rightarrow \mathbf{R}^d$ creates at each time step $d$ features, such that, $x^u = g(id(u))$ which corresponds to engineered features for the customer using historical data from previous calls to the center, as well as using other information (e.g., demographics).

We consider a learning model (multi-class classifier) $f: \mathbf{R}^d \rightarrow \{0,1, ..., N\}$, such that, $\hat{y}^u = f(x^u)$ where $N$ is the number of departments, and $\hat{y}^u$ is the predicted department to be transferred to. Note that the time taken for feature extraction and model prediction is negligible (i.e., in the order of milliseconds). At this point, the user receives a message to confirm whether or not the prediction is correct (i.e., in the order of a few seconds). Let us assume that the average time required is $t_{stage^{pred}}.$

The main objective is to minimize the average time duration of each call $T^{c}$ and, as a result, the total average duration time $T^{total}$, as defined in Eqs \ref{eq:userTime} and \ref{eq:totalTime}. We define the time duration of a call $c$ for which the model provided a correct prediction as follows:
\begin{equation}
\begin{aligned}
  T^{c}_{correct} &= t_{stage^{pred}} + t_{stage^2} \\
   &= t_{stage^{pred}} + (T^c - t_{stage^1} )
\end{aligned}
\end{equation}
\noindent where $T^c$ is the original average time taken without any prediction as defined in Eq. \ref{eq:userTime}. In this case $T^{c}_{correct} << T^c$ as the stage 1 is by-passed which, typically, requires tens of seconds.

Similarly, let us define the average time duration of a call for which the model provided an incorrect prediction as shown below. In this case $T^{c}_{incorrect} > T^c$.
\begin{equation}
   T^{c}_{incorrect} = t_{stage^{pred}} + T^c
\end{equation}

Based on the number of calls which correspond to correct and incorrect predictions respectively, the total average duration time is defined as:
\begin{equation}\label{eq:totalTimePred}
   T^{total}_{pred} = \sum_{c=1}^{T_P+T_N} T^{c}_{correct} + \sum_{c=1}^{F_P+F_N} T^{c}_{incorrect}
\end{equation}
\noindent where $T_P, T_N, F_P$ and $F_N$ correspond to the number of True Positives, True Negatives, False Positives, and False Negatives respectively.

Thus far we have considered the general case where a multi-class classifier is used. For completeness, when the task is binary, that is, to predict whether a user is calling for a particular service $SERVICE\_A$ or for any other service $OTHER$, the previous equation is replaced with the following:

\begin{equation}\label{eq:totalTimePredBinary}
   T^{total}_{pred} = \sum_{c=1}^{T_P} T^{c}_{correct} + \sum_{c=1}^{F_P} T^{c}_{incorrect} + \sum_{c=1}^{T_N+F_N} T^{c}
\end{equation}

\section{Experimental Setup}\label{sec:setup}

\subsection{Case study}
We have collaborated with one of the largest telecommunications and Internet provider in the country, to provide a proof-of-concept to our proposed ML-based customer - agent pairing. The objective of this initial case study is to predict whether a user calls to purchase a specific service $SERVICE\_A$, otherwise she / he calls for another reason $OTHER$. To achieve this, we extract the following types of information related to each user.

\textbf{Demographics:} It includes general demographic information, such as, age, language, address, and the user type (e.g., company or individual).

\textbf{Customer Profile:} It contains detailed information about each service or product the customer has used or has been using (e.g., start date, expiration date, description, and status).

\textbf{Customer Interaction:} It includes information about issues that were reported by a user in the past via the Customer Call Center or at a retail store; this source includes information, such as, date and time, and department handled by.

Feature engineering is performed on the extracted features related to customer profile and interaction, for example, "number of previous calls in the last three months for $SERVICE\_A$".

After the dataset creation, we split the dataset into training, validation, and test sets as shown in Table \ref{splits}.

\begin{table}[h]
\caption{\label{splits} Dataset description}
\centering
\begin{tabular}[c]{| p{2cm} | p{2cm} | p{2cm} | p{2cm} | p{2cm} | p{2cm} |} 
\hline
 \multicolumn{1}{|c|}{\textbf{Dataset}} & \multicolumn{1}{c|}{\textbf{Duration (months)}} &\multicolumn{1}{c|}{\textbf{Unique Callers}} &\multicolumn{1}{c|}{\textbf{Service\_A Calls}} &\multicolumn{1}{c|}{\textbf{Other Calls}} &\multicolumn{1}{c|}{\textbf{Total Calls}} \\
\hline
   Train &12 &162563 &18581 &822102 &849683 \\[0.9ex]
   Validation &2 &56317 &2289 &118317 &120606 \\[0.9ex]
   Test &1 &34128 &1378 &59407 &60785 \\[0.9ex]
\hline
\end{tabular}
\label{table:splits}
\end{table}

\subsection{Compared methods}

\textbf{Rule-based}: Combination of rules derived based on domain knowledge, as well as, extensive analysis of historical data.

\textbf{Logistic Regression (LR)} \cite{cox1958regression}: It is a statistical algorithm that models the probability of an event taking place by analyzing the linear relationship between one or more existing independent variables.

\textbf{Decision Tree (DT)} \cite{Wu2007Top}: It is a classification algorithm that predicts the class of a target variable by learning decision rules inferred from prior data.

\textbf{Random Forest (RF)} \cite{bishop2006} \cite{breiman2001random}:
It is a tree-based, ensemble learning algorithm, i.e., it depends on multiple tree-based learners which make individual predictions that are then averaged together.

\textbf{Extreme Gradient Boosting (XGBoost)} \cite{chen2016xgboost}:
It is a machine learning technique that produces a prediction model in the form of an ensemble of weak prediction models, which are typically tree-based. This technique builds a model in a stage-wise fashion and combines weak learners into a single strong learner. As each weak learner is added, a new model is fitted to provide a more accurate estimation. The XGBoost classifier is a tree-based ensemble machine learning algorithm with Gradient Boosting as its main component.

\textbf{Multilayer Perception (MLP)} \cite{bishop2006}:
It is a feed-forward neural network that consists of an input and an output layer, and can have multiple hidden layers. MLP uses the backpropagation algorithm for training which computes the gradient of the loss function with respect to the weights of the neural network.

\subsection{Evaluation metrics}
We analyze the results of the classifier using a \textbf{confusion matrix}. The confusion matrix classifies the results into True Positives ($T_P$), True Negatives ($T_N$), False Positives ($F_P$), and False Negatives ($F_N$). 

Classifiers are typically evaluated using the accuracy metric. However, this metric becomes unsuitable as it is biased towards the majority class. In this study we adopt two widely accepted metrics which are less sensitive to imbalance; these are, F1-score (F1) \cite{he2009learning} and Geometric Mean (GM).

\textbf{F1-score (F1)} is defined as the harmonic mean of Precision and Recall. Specifically, Precision ($P$) provides information concerning the rate at which the algorithm detects $SERVICE\_A$ over all detection of $SERVICE\_A$ given by: 
 
\begin{equation}\label{eq:P}
P= \frac{T_P}{T_P + F_P}.
\end{equation}

Similarly, Recall ($R$) is the ratio at which the algorithm detects $SERVICE\_A$ over all possible $SERVICE\_A$ given by:

\begin{equation}\label{eq:R}
R= \frac{T_P}{T_P + F_N}.
\end{equation}

Finally, the $F1$-score ($F1$) is the weighted average of $P$ and $R$ and is the measure of accuracy on the data set given by

\begin{equation}\label{eq:F1}
F1=2\left(\frac{P\times R}{P+R}\right).
\end{equation}

The $F1$-score gets a higher value (near $1$) when the $F_P$ and $F_N$ are low. If a system is performing poorly by generating more $F_P$ and $F_N$, the $F1$-score will be low (near $0$).

\textbf{Geometric Mean (GM)} is defined as the geometric mean of Recall and Specificity, and is given by 

\begin{equation}\label{eq:gm}
GM=\sqrt{R \times S},
\end{equation}
where the Specificity ($S$), which is defined as the true negative rate, is given by 
\begin{equation}
S={\frac{T_N}{T_N + F_P}}.
\end{equation}

Note that GM has the desirable property of being high when both $R$ and $S$ are high, and when their difference is small \cite{he2009learning}. For this reason, we introduce a combined metric which is defined as the geometric mean of $F1$ and $GM$:
\begin{equation}\label{eq:gm}
F1-GM=\sqrt{F1 \times GM}
\end{equation}

\section{Experimental Results}\label{sec:results}

\subsection{Role of the learning model}

In this section we explore the performance of each model on the validation set. Due to the stochastic nature of the learning models we repeat the experiment 20 times, and report the average performance. An overview of the results can be seen in Table \ref{MLValRes}. The MLP appears to be the best performing model based on the combined $F1-GM$ metric; specifically, it produces the highest $F1$ and the second highest $GM$.

\begin{table}[h]
\caption{\label{MLValRes} Performance of machine learning models on the validation set.}
\centering
\begin{tabular}[c]{| p{2cm} | p{2cm} | p{2cm} | p{2cm} | } 
\hline
\multicolumn{1}{|c|}{\textbf{Model}} & \multicolumn{1}{|c|}{\textbf{F1}} & \multicolumn{1}{|c|}{\textbf{GM}} & \multicolumn{1}{|c|}{\textbf{F1-GM}} \\
\hline
  LR &47.69 &75.93 &60.17\\[0.9ex]
  DT &46.74 &71.34 &57.74 \\[0.9ex]
  RF &47.31 &\textbf{77.89} &60.70\\[0.9ex]
  XGBoost &50.00 &74.25 &60.93\\[0.9ex]
  MLP & \textbf{50.17}& 76.39& \textbf{61.91}\\[0.9ex]
\hline
\end{tabular}
\label{table:MLValRes}
\end{table}

\subsection{Comparative study}
In this section we compare the machine learning algorithm to the manual rules method. We have selected the highest performing machine learning algorithm, MLP and the best rule combination for manual rules method. Table \ref{TestRes} presents the average performance of MLP and manual rules on the test set. The proposed leaning-based method yields an $F1$ and $GM-F1$ scores which are 4.5 and 2.15 times better than the rule-based method. This significant improvement can reduce the waiting times, thus enhancing the customer experience, as well as it allows an organisation to allocate its limited resources more efficiently.

\begin{table}[h]
\caption{\label{TestRes} Comparative study on the test set.}
\centering
\begin{tabular}[c]{| p{2cm} | p{2cm} | p{2cm} | p{2cm} | } 
\hline
\multicolumn{1}{|c|}{\textbf{Algorithm}} & \multicolumn{1}{|c|}{\textbf{F1}} & \multicolumn{1}{|c|}{\textbf{GM}} & \multicolumn{1}{|c|}{\textbf{F1-GM}} \\
\hline
  Manual Rules &10.67 &74.04 &28.10\\[0.9ex]
  MLP & \textbf{48.55}& \textbf{75.39}& \textbf{60.50}\\[0.9ex]
\hline
\end{tabular}
\label{table:TestRes}
\end{table}

\subsection{Empirical analysis of the results}
Table \ref{Conf} presents the confusion matrix of Manual Rules and MLP on the test set. In this section, given the numbers presented in Table \ref{Conf} and the formulated equations in section 3, we analyse three methods. The first one is the traditional way in which a call centre operates, i.e., without any prediction method in-place. The second and third methods are the rule-based and ML-based respectively. MLP achieves a significant improvement over the rule-based method; we notice here the huge difference in the number of true negatives and false positives.

\begin{table}[h]
\caption{\label{Conf}Confusion matrix of Manual Rules and MLP algorithms on the test set.}  
\centering
\begin{tabular}[c]{| p{2.5cm} | p{1.2cm} | p{1.2cm} | p{1.2cm} | p{1.2cm} |p{1.2cm} |} 
\hline
\multicolumn{1}{|c|}{\textbf{Algorithm}} &
\multicolumn{1}{|c|}{$\mathbf{T_N}$} &
\multicolumn{1}{|c|}{$\mathbf{F_P}$} &
\multicolumn{1}{|c|}{$\mathbf{F_N}$} &
\multicolumn{1}{|c|}{$\mathbf{T_P}$} \\
\hline
  Manual Rules & 41675 & 17732 & 301 &  1077 \\[0.9ex]
  MLP & 58308 & 1099 & 579 &  799 \\[0.9ex]
\hline
\end{tabular}
\label{table:Conf}
\end{table}

\textbf{Traditional (no prediction)}.
Let us now consider the traditional method of not having any prediction method in-place. Given equation \ref{eq:totalTime} and the total number of calls which is $60785$, the total time duration equals to:
\begin{equation}
\begin{aligned}
   T^{total}_{trad} &= \sum_{c=1}^{60785} T^c\\
   &= 60785t_{stage^1} + 60785t_{stage^2}
\end{aligned}
\end{equation}

\textbf{Rule-based}. Based on Eq. \ref{eq:totalTimePredBinary}, the total duration time is:

\begin{equation}
\begin{aligned}
   T^{total}_{pred\_rules} &= \sum_{c=1}^{1077} T^{c}_{correct} + \sum_{c=1}^{17732} T^{c   }_{incorrect} + \sum_{c=1}^{41976} T^{c}_{other} \\
   &= 18809t_{stage^{pred}}+59708t_{stage^{1}}+60785t_{t_{stage^{2}}} \\
   &= 18809t_{stage^{pred}}+59708t_{stage^{1}}+877t_{stage^{1}}-877t_{stage^{1}}+60785t_{t_{stage^{2}}}\\
   &= T^{{total}_{trad}} + 18809t_{stage^{pred}} - 877t_{stage^{1}}
\end{aligned}
\end{equation}

\textbf{MLP}. Based on Eq. \ref{eq:totalTimePredBinary}, the total duration time is:

\begin{equation}
\begin{aligned}
   T^{total}_{pred\_MLP} &= \sum_{c=1}^{799} T^{c}_{correct} + \sum_{c=1}^{1099} T^{c}_{incorrect} + \sum_{c=1}^{58887} T^{c}_{other}\\
   &= 1898t_{stage^{pred}}+59986t_{stage^1}+60785t_{stage^2}\\
   &= 1898t_{stage^{pred}}+59986t_{stage^1}+799_{stage^1}-799_{stage^1}+60785t_{stage^2}\\
   &= T^{total}_{trad} + 1898t_{stage^{pred}} - 799_{stage^1}
\end{aligned}
\end{equation}

We can further derive that the average time required by the traditional method is larger than that of the proposed method if this condition is true:

\begin{equation}
\begin{aligned}
   &T^{total}_{trad} > T^{total}_{pred\_MLP}\\
   &\Rightarrow 799t_{stage^{1}} - 1898t_{stage^{pred}} > 0 \\
   &\Rightarrow t_{stage^{1}} > 2.38t_{stage^{pred}}
\end{aligned}
\end{equation}

Similarly, the average time required by the rule-based method is larger than that of the proposed method if the following condition is true:
\begin{equation}
\begin{aligned}
   &T^{total}_{pred\_rules} > T^{total}_{pred\_MLP}\\
   &\Rightarrow 16911t_{stage^{pred}} - 78t_{stage^{1}} > 0 \\
   &\Rightarrow t_{stage^{1}} < 216.81t_{stage^{pred}}
\end{aligned}
\end{equation}

In our case study, reasonable values for the average duration of the prediction and first stages are $t_{stage^{pred}} = 5 sec$ and $t_{stage^{1}} = 45 sec$ respectively. Both the above conditions are met.
Specifically, the proposed ML-based method reduces the total average waiting time in the test set (1 month) compared to the rule-based method by more than 22 hours (= $T^{total}_{pred\_rules} - T^{total}_{pred\_MLP}$), and to the traditional method by more than 7 hours (= $T^{total}_{trad} - T^{total}_{pred\_MLP}$).

\section{Conclusions and Future Work}\label{sec:conclusions}
We have collaborated with the largest provider of telecommunications and Internet access in the country, and we have formulated the customer-agent pairing problem as a machine learning problem. The proposed learning-based method causes a significant improvement in performance of about 215\% (i.e., 2.15 times better) compared to a rule-based method. One future direction is to examine the effect of methods, such as, cost-sensitive learning and resampling, to address the class imbalance problem, in an attempt to further improve our results.

\subsubsection{Acknowledgements} This work has been supported by the CYTA-KIOS Research Collaboration Agreement, the European Union Horizon 2020 program under Grant Agreement No. 739551 (KIOS CoE), and the Government of the Republic of Cyprus through the Deputy Ministry of Research, Innovation and Digital Policy.

\bibliographystyle{splncs04}
\bibliography{mybib}

\end{document}